\documentclass[10pt,twocolumn,letterpaper]{article}

\usepackage{cvpr}
\usepackage{times}
\usepackage{epsfig}
\usepackage{graphicx}
\usepackage{amsmath}
\usepackage{amssymb}
\usepackage{dsfont}
\usepackage[outline]{contour}

\usepackage{xfrac}
\usepackage{subcaption}
\usepackage[export]{adjustbox}
\usepackage{rotating}
\usepackage{array}
\usepackage{siunitx}
\usepackage{units}

\definecolor{myblue}{rgb}{.1,0.3,0.9}
\definecolor{rowblue}{RGB}{220,230,240}
\definecolor{barorange}{RGB}{255,166,0}
\definecolor{barblue}{RGB}{88,80,141}
\definecolor{barblue2}{RGB}{0,63,92}
\usepackage{booktabs}
\usepackage{tablefootnote}
\usepackage{colortbl}
\usepackage{tabularx}
\usepackage{multirow}

\usepackage[pagebackref=true,breaklinks=true,letterpaper=true,colorlinks,bookmarks=false]{hyperref}

\cvprfinalcopy 

\def\cvprPaperID{5231} 

\makeatletter
\renewcommand*{\@fnsymbol}[1]{\ifcase#1\or\else\@arabic{\numexpr#1-1\relax}\fi}
\makeatother

\ifcvprfinal\fi

\definecolor{Gray}{rgb}{0.5,0.5,0.5}
\definecolor{darkblue}{rgb}{0,0,0.7}
\definecolor{orange}{rgb}{1,.5,0} 
\definecolor{red}{rgb}{1,0,0} 

\newcommand{\yaw}{\phi} 
 
\newcommand{\clight}{c} 				
\newcommand{\radvel}{v_r}
\newcommand{\phaseshift}{\Omega}		
\newcommand{\fbeat}{f_\text{beat}} 
\newcommand{\fdoppler}{f_\text{Doppler}} 
\newcommand{\pointcloud}{\tilde{\mathbf{U}}}

\newcommand{\Refl}{\mathit{\rho}}             

\newcommand{\wallpos}{\Vect{w}}               
\newcommand{\voxelpos}{\Vect{x}}      
\newcommand{\virtualpos}{\Vect{x^{\prime}}}    
\newcommand{\wallnormal}{\Vect{n_w}}            
\newcommand{\pixelpos}{\Vect{c}}               
\newcommand{\wall}{\Vect{p}}

\newcommand{\outgoingdir}{\Vect{\omega_o}}
\newcommand{\incidentdir}{\Vect{\omega_i}}
\newcommand{\chirptime}{T_\text{tot}}
\newcommand{\risetime}{T_m}

\newcommand{\tx}{\Vect{l}}
\newcommand{\rx}{\Vect{c}}

\newcommand{\att}{\alpha}               

\newcommand{\Vect}[1]   {{\ensuremath{\mathbf{\lowercase{#1}}}}} 

\newcommand{\cross}      {\times}

\newcommand*{\norm}[1]{\left\lVert#1\right\rVert}		
\newcommand*{\abs}[1]{\left\lvert#1\right\rvert}		
\DeclareMathOperator*{\sign}{sgn}						

\begin{document}

\title{\vspace{-35pt}Seeing Around Street Corners:\vspace{4pt}\\ Non-Line-of-Sight Detection and Tracking In-the-Wild Using Doppler Radar\vspace{-10pt}}

\author{Nicolas Scheiner\thanks{\textsuperscript{*}Equal contribution.}\textsuperscript{*1}\hspace{0.1in}\vspace{-15pt}\\
\and
Florian Kraus\textsuperscript{*1}\hspace{0.1in}\vspace{-3pt}\\
\and
Fangyin Wei\textsuperscript{*2}\hspace{0.1in}\vspace{-3pt}\\
\and
Buu Phan\textsuperscript{3}\hspace{0.3in}\vspace{-3pt}\\
\and
Fahim Mannan\textsuperscript{3}\hspace{0.2in}\vspace{-3pt}\\
\and
Nils Appenrodt\textsuperscript{1}\hspace{0.05in}\vspace{-3pt}\\
\and
Werner Ritter\hspace{0.2mm}\textsuperscript{1}\hspace{-0.2in}\vspace{-3pt}\\
\and
J{\"u}rgen Dickmann\textsuperscript{1}\hspace{0.0in}\vspace{-3pt}\\
\and
Klaus Dietmayer\hspace{0.3mm}\textsuperscript{4}\hspace{0.1in}\vspace{-2pt}\\
\and
Bernhard Sick\hspace{0.2mm}\textsuperscript{5}\hspace{0.1in}\vspace{-2pt}\\
\and
Felix Heide\hspace{0.2mm}\textsuperscript{2,3}\hspace{0.0in}\vspace{-2pt}\\~\vspace{-10pt}
\and
\textsuperscript{1}Mercedes-Benz AG\hspace{0.15in}
\textsuperscript{2}Princeton University\hspace{0.15in}
\textsuperscript{3}Algolux\hspace{0.15in}
\textsuperscript{4}Ulm University\hspace{0.15in}
\textsuperscript{5}University of Kassel
\vspace{-0.2cm}
}

\maketitle
\thispagestyle{plain}
\pagestyle{plain}

\begin{abstract}\vspace{-0.3cm}
Conventional sensor systems record information about directly visible objects, whereas occluded scene components are considered lost in the measurement process. Non-line-of-sight (NLOS) methods try to recover such hidden objects from their indirect reflections -- faint signal components, traditionally treated as measurement noise. Existing NLOS approaches struggle to record these low-signal components outside the lab, and do not scale to large-scale outdoor scenes and high-speed motion, typical in automotive scenarios. In particular, optical NLOS capture is fundamentally limited by the quartic intensity falloff of diffuse indirect reflections.
In this work, we depart from visible-wavelength approaches and demonstrate detection, classification, and tracking of hidden objects in large-scale dynamic environments using Doppler radars that can be manufactured at low-cost in series production.
To untangle noisy indirect and direct reflections, we learn from temporal sequences of Doppler velocity and position measurements, which we fuse in a joint NLOS detection and tracking network over time. We validate the approach on in-the-wild automotive scenes, including sequences of parked cars or house facades as relay surfaces, and demonstrate low-cost, real-time NLOS in dynamic automotive environments.
\end{abstract}

\vspace{-4mm}
\section{Introduction}
Conventional sensor systems capture objects in their direct line of sight, and, as such, existing computer vision methods are capable of detecting and tracking only the visible scene parts~\cite{girshick2014rich, he2015spatial, ren2015faster, redmon2016you, chen2017multi,ku2018joint,zhou2018voxelnet,luo2018fast}, whereas occluded scene components are deemed lost in the measurement process. Non-line-of-sight (NLOS) methods aim at recovering information about these occluded objects from their indirect reflections or shadows on visible scene surfaces, which are again in the line of sight of the detector.
While performing scene understanding of occluded objects may enable applications across domains, including remote sensing or medical imaging, especially autonomous driving applications may benefit from detecting approaching traffic participants that are occluded.
\begin{figure}[t!]
\vspace{-2pt}
    \centering
		\includegraphics[width=1.0\linewidth]{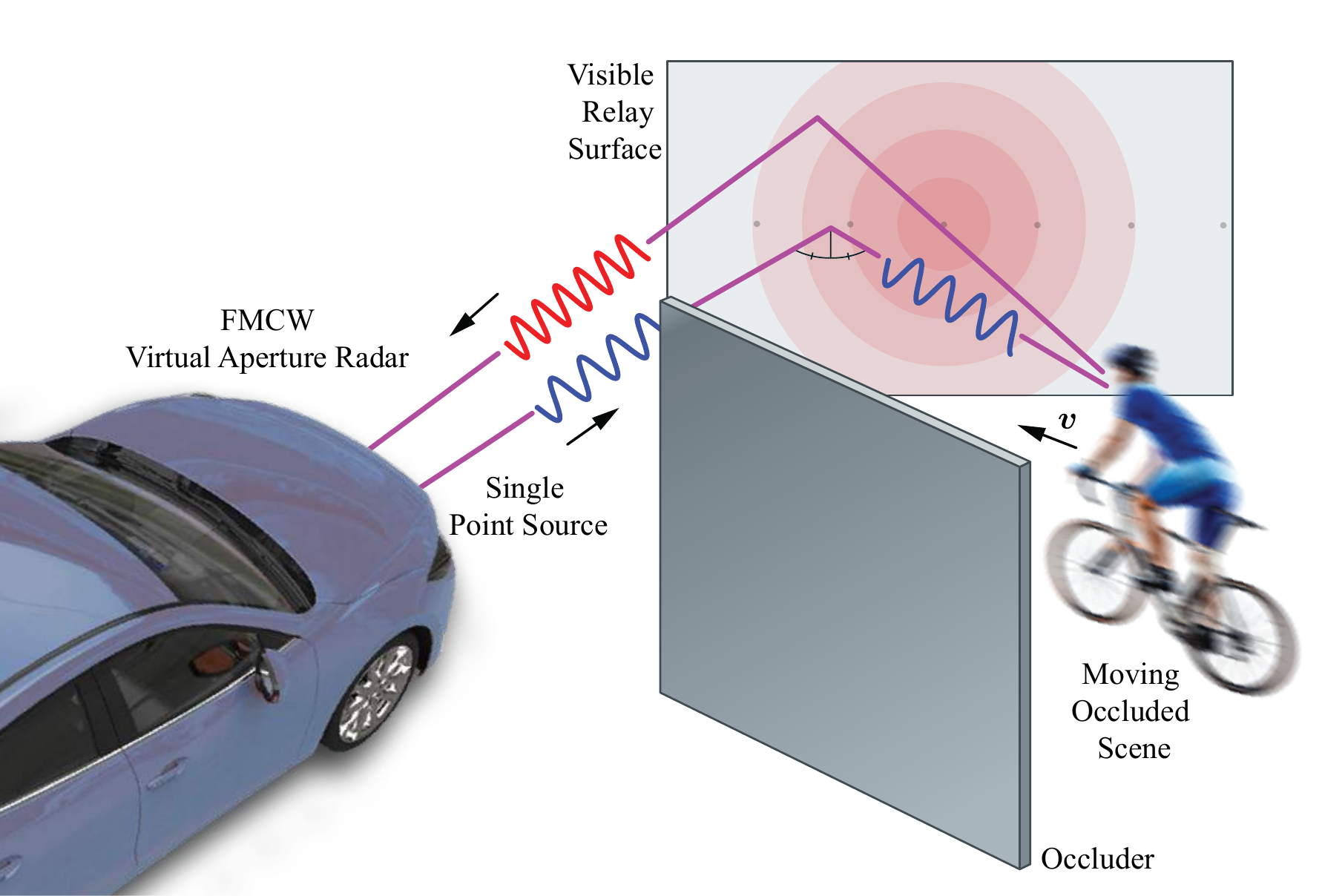}
		\vspace{-6pt}
    \caption{We demonstrate that it is possible to recover moving objects outside the direct line of sight in large automotive environments from Doppler radar measurements. Using static building facades or parked vehicles as relay walls, we jointly classify, reconstruct, and track occluded objects.}
    \label{fig:Teaser}
    \vspace{-10pt}
\end{figure}

Existing NLOS imaging methods struggle outside controlled lab environments, and they struggle with large-scale outdoor scenes and high-speed motion, such as in typical automotive scenarios. The most successful NLOS imaging methods send out ultra-short pulses of light and measure their time-resolved returns~\cite{Velten:2012recovering,Pandharkar:2011,Gupta:12,buttafava2015non,tsai2017geometry,arellano2017fast,o2018confocal,liu2019phasor_nlos}. In contrast to a conventional camera, such measurements allow existing methods to unmix light paths based on their travel time~\cite{Abramson:78:LIF,Kirmani:2009,Naik:2011,Pandharkar:2011}, effectively trading angular with temporal resolution.
As a result, pulse widths and detection at a time scale of $<$ \SI{10}{\pico\second} is required for room-sized scenes, mandating specialized equipment which suffers from low photon efficiency, high cost, and slow mechanical scanning. As intensity decreases quartically with the distance to the visible relay wall, current NLOS methods are limited to meter-sized scenes even when exceeding the eye-safety limits for a Class~1 laser (\eg Velodyne HDL-64E) by a factor of 1000~\cite{Lindell:2019:Wave}. Moreover, these methods are impractical for dynamic scenes as scanning and reconstruction takes up minutes~\cite{liu2019phasor_nlos,arellano2017fast}. Unfortunately, alternative approaches based on amplitude-modulated time-of-flight sensors~\cite{heide2014diffuse,kadambi2016occluded,kadambi2013coded} suffer from modulation bandwidth limitations and ambient illumination~\cite{lange00tof}, and intensity-only methods~\cite{chen2019steady,saunders2019computational,bouman2017turning} require highly reflective objects. Large outdoor scenes and highly dynamic environments remain an open challenge.

In this work, we demonstrate that it is possible to detect and track objects in large-scale dynamic scenes outside of the direct line-of-sight using automotive Doppler radar sensors, see Fig.~\ref{fig:Teaser}. Departing from visible-wavelength NLOS approaches which rely on diffuse indirect reflections on the relay wall, we exploit the fact that specular reflections dominate on the relay wall for mm-wave radar signals, \ie, when the structure size is an order of magnitude larger than the wavelength. As such, in contrast to optical NLOS techniques, phased array antenna radar measurements preserve the angular resolution and emitted radio frequency (RF) power in an indirect reflection, which enables us to achieve longer ranges. Conversely, separating direct and indirect reflections becomes a challenge. To this end, we recover indirectly visible objects relying on their Doppler signature, effectively suppressing static objects, and we propose a joint NLOS detection and tracking network, which fuses estimated and measured NLOS velocity over time. We train the network in an automated fashion, capturing training labels along with data with a separate positioning system, and validate the proposed method on a large set of automotive scenarios. By using facades and parked cars as reflectors, we show a first application of NLOS collision warning at urban intersections.
	
Specifically, we make the following contributions:
\begin{itemize}
	\setlength\itemsep{.2em}
	\item We formulate an image formation model for Doppler radar NLOS measurements. Based on this model, we derive the position and velocity of an occluded object.
	\item We propose a joint NLOS detection and tracking network, which fuses estimated and measured NLOS velocity over time. For occluded object labeling, we acquire our data with an automated positioning system.
	\item We validate our system on in-the-wild automotive scenarios, and as a first application of this new imaging modality, demonstrate collision warning for vulnerable road users before seeing them in direct line of sight.
	\item The experimental training and validation data sets and models will be published\footnotemark.
\end{itemize}

\section{Related Work}
\vspace{0em}\noindent\textbf{Optical Non-Line-of-Sight Imaging}
A growing body of work explores optical NLOS imaging techniques~\cite{Pandharkar:2011,Velten:2012recovering,Gupta:12,kadambi2016occluded,o2018confocal,tsai2017geometry,arellano2017fast,pediredla2017reconstructing,Xu:18,liu2019phasor_nlos}. Following Kirmani et al.~\cite{Kirmani:2009}, who first proposed the concept of recovering occluded objects from time-resolved light transport, these methods directly sample the temporal impulse response of a scene by sending out pulses of light and capturing their response using detectors with high temporal precision of $<$~\SI{10}{\pico\second}, during which the pulses travel a distance of \SI{3}{\milli\meter}. While early work relies on costly and complicated streak camera setups~\cite{Velten:2012recovering,Velten:2012:Visualizing}, a recent line of work uses single photon avalanche diodes (SPAD)~\cite{buttafava2015non,o2018confocal,liu2019phasor_nlos}. 
Katz et al.~\cite{katz2012looking,katz2014non} demonstrate that correlations in the carrier wave itself can be used to realize fast single shot NLOS imaging, however, limited to scenes at microscopic scales~\cite{katz2014non}.

\vspace{0.4em}\noindent\textbf{Non-Line-of-Sight Tracking and Classification}
Several recent works use conventional intensity images for NLOS tracking and localization~\cite{klein2016tracking,caramazza2018neural,chan2017non,bouman2017turning,chen2019steady}. The ill-posedness of the underlying inverse problem limits these methods to localization with highly reflective targets~\cite{bouman2017turning,chen2019steady}, sparse dark background, or only scenes with additional occluders present~\cite{saunders2019computational,bouman2017turning}. Unfortunately, recent acoustic methods~\cite{lindell2019acoustic} are currently limited to meter-sized lab scenes and minutes of acquisition time. All of these existing methods have in common that they are impractical for large and dynamic outdoor environments.

\vspace{0.4em}\noindent\textbf{Radio Frequency Non-Line-of-Sight Imaging}
A further line of work has explored imaging, tracking, and pose estimation through walls using RF signals~\cite{adib2015capturing,adib20143d,adib2013see,richards2010principles,wilson2009throughwall,zhao2018through}. However, RF signals are not reflected when traveling through typical interior wall material, such as drywall, drastically simplifying through-the-wall vision tasks. As a result, only a few works have explored NLOS radar imaging and tracking~\cite{sume2011radar,rabaste2017around,zetik2015looking}. These methods backpropagate raytraced high-order-bounce signals, which requires scenes with multiple known (although they are occluded) hidden relay walls. For the in-the-wild scenarios tackled in this work without prior scene knowledge, only third-bounce measurements, and with imperfect relay walls, \eg, a parked sequence of vehicles, these methods are impractical. Moreover, traditional filtering and backprojection estimation suffers from large ambiguities at more than \SI{10}{\meter} in the presence of realistic measurement noise~\cite{rabaste2017around}. In this work, we address this challenge with a data-driven joint detection and tracking method, allowing us to demonstrate practical NLOS detection in-the-wild using radar systems which have the potential for low-cost mass production in the near future.
\footnotetext{\href{https://github.com/princeton-computational-imaging/doppler\_nlos}{https://github.com/princeton-computational-imaging/doppler\_nlos} for code and models.}

\vspace{-1mm}
\section{Observation Model}
\vspace{-1mm}
\label{sec:image_formation}
When a radar signal gets reflected off a visible wall onto a hidden object, some of the signal is scattered and reflected back to the wall where it can be observed, see Fig.~\ref{fig:supp_array}. Next, we derive a forward model including such observations.
\subsection{Non-Line-of-Sight FMCW Radar}
\begin{figure}[t!] 
	\centering
	\vspace{-7pt}
	\includegraphics[width=0.85\linewidth]{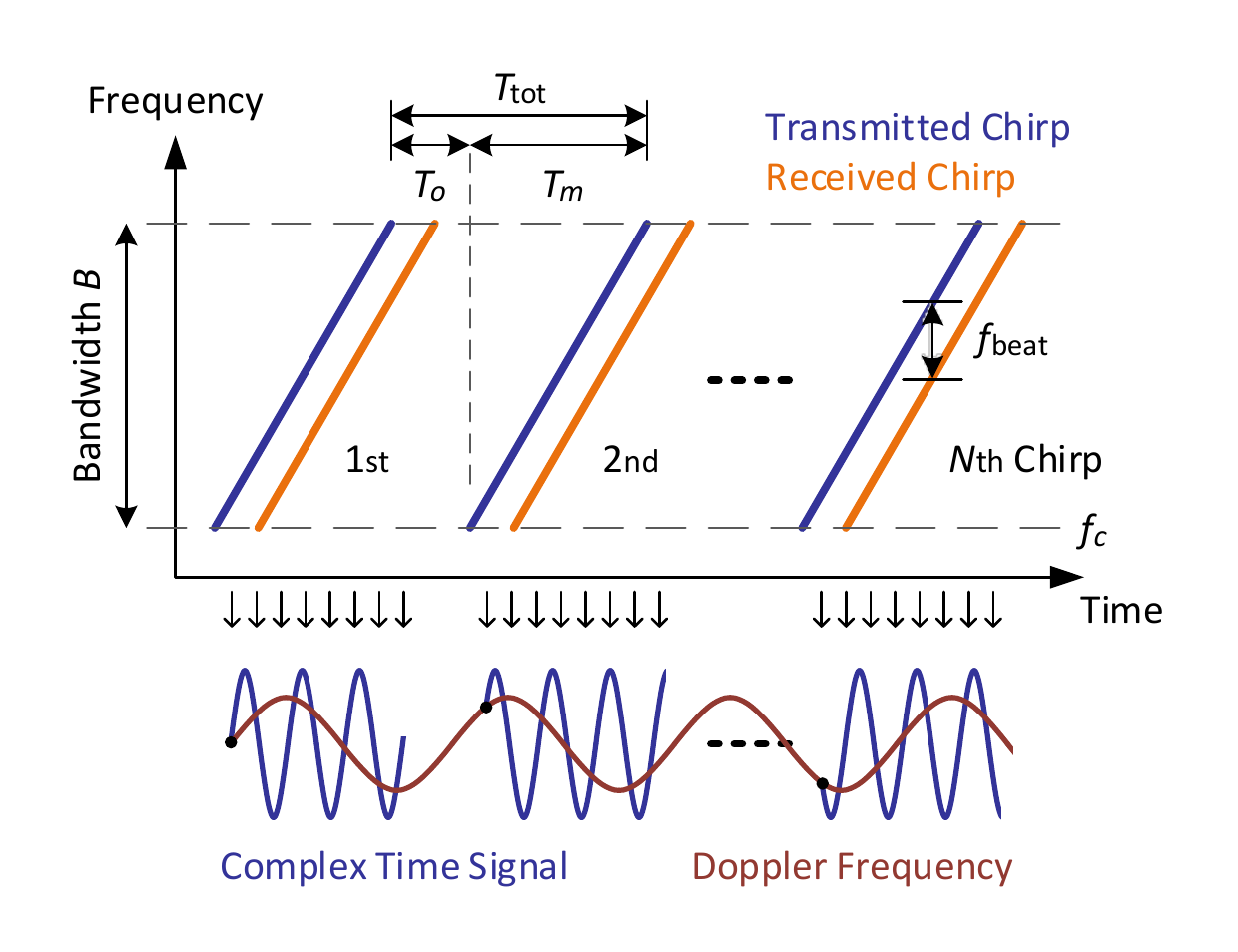}
	\vspace{-7pt}
    \caption{Chirp sequence modulation principle for a single receiver-transmitter antenna: $N$ consecutive frequency ramps are sent and received with a frequency shift $\fbeat$ corresponding to the distance of the reflector. Each frequency ramp is sampled and the phase of the received signal is estimated at each chirp and range bin. The phase shift between consecutive chirps corresponds to the Doppler frequency.}
		\vspace{-4pt}
    \label{fig:supp_chirp}
		\vspace{-4pt}
\end{figure}

Radar sensors emit electromagnetic (EM) waves, traveling at the speed of light $\clight$, which are reflected by the scene and received by the radar sensors.
In this work, we use a frequency-modulated continuous-wave (FMCW) Doppler radar with multiple input multiple output (MIMO) array configuration, which can resolve targets in range $r$, azimuthal angle $\phi$, and radial Doppler velocity $\radvel$.
Instead of a single sinusoidal EM wave, our FMCW radar sends out linear frequency sweeps~\cite{brooker2005understanding} over a frequency band $B$ starting from the carrier frequency $f_c$, that is
\begin{equation}\label{eq:emitted_signal}
g(t) = \cos \left( 2\pi f_c t + \pi \frac{B}{\risetime} t^2 \right),
\end{equation}
with $\risetime$ being the sweep rise time.
The instantaneous frequency of this signal is $\nicefrac{1}{2\pi} \, \nicefrac{d}{dt}\left(2\pi f_c t + \pi \, \nicefrac{B}{\risetime} t^2\right) = f_c + \nicefrac{B}{\risetime} t$, that is a linear sweep varying from $f_c$ to $f_c + B$ with slope $\nicefrac{B}{\risetime}$, which is plotted in Fig.~\ref{fig:supp_chirp}.

\begin{figure}[t!] 
	\centering
	\vspace{-12pt}
	\includegraphics[width=.81\linewidth]{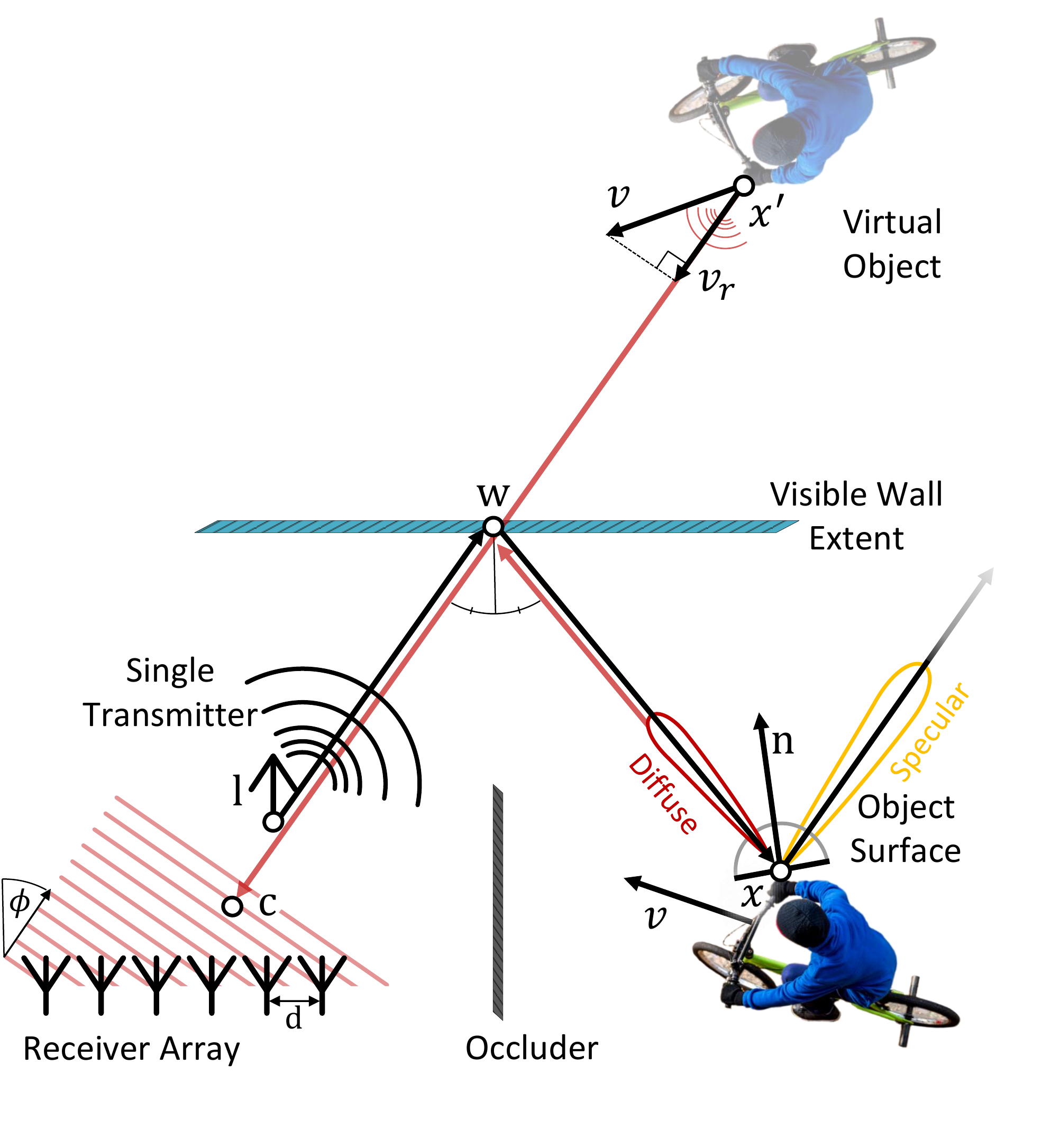} 
	\captionsetup{skip=-10pt}
	\caption{Radar NLOS observation. For mm-wavelengths, typical walls appear flat, and reflect radar waves specularly. We measure distance, angle and Doppler velocity of the indirect diffuse backscatter of an occluded object to recover its velocity, class, shape, and location.}
	\vspace{-10pt}
	\label{fig:supp_array}
\end{figure}

The emitted signal $g$ propagates through the visible and occluded parts of the scene, that is, this signal is convolved with the scene's impulse response. For a given emitter position $\tx$ and receiver position $\rx$ the received signal becomes\vspace{-2pt}
\begin{small}\begin{align}
\label{eq:transport_model}
	& s(t,\rx,\tx,\wallpos)  =  \hspace{-3pt} \int_{\Lambda} \, \att(\voxelpos) \, \Refl \left( \voxelpos - \wallpos, \wallpos - \voxelpos \right)  \, \cdot \\ 
	& \! \frac{1}{(r_{\tx\wallpos} \! + \!  r_{\voxelpos\wallpos})^2} \, \frac{1}{(r_{\voxelpos\wallpos} \! + \! r_{\wallpos\rx})^2} \,\, g\!\left(\! t  -  \frac{r_{\tx\wallpos} \! + \! 2 \, r_{\voxelpos\wallpos} \! + \! r_{\wallpos\rx}}{\clight} \!\right) d \sigma(\voxelpos), \nonumber
\end{align}\noindent\ignorespacesafterend
\parskip=1pt
\end{small}
see Fig.~\ref{fig:supp_array}, with $\wallpos$ and $\voxelpos$ being the positions on the relay wall and the object surface $\Lambda$, the surface measure $\sigma$ on $\Lambda$, $\att$ as the albedo, and $\Refl$ denoting the bi-directional reflectance distribution function (BRDF), which depends on the incident direction $\incidentdir=\voxelpos - \wallpos$ and outgoing direction $\outgoingdir=\wallpos - \voxelpos$. The distance $r$ describes here the distance between the subscript positions, and its squared inverse in Eq.~\eqref{eq:transport_model} models the intensity falloff due to spherical travel, which we approximate as not damped by the specular wall, and diffuse backscatter from object back to the receiver $\rx$.

\vspace{0.8em}\noindent\textbf{Reflection Model}
The scattering behavior $\Refl$ depends on the surface properties.
Surfaces that are flat, relative to the wavelength $\lambda$ of $\approx$~\SI{5}{\milli\meter} for typical \SI{76}{GHz}-\SI{81}{GHz} automotive radars, will result in a specular response.
As a result, the transport in Eq.~\eqref{eq:transport_model} treats the relay wall as a mirror, see Fig.~\ref{fig:supp_array}.
We model the reflectance of the hidden and directly visible targets following~\cite{chen2019steady} with a diffuse and specular term as
\vspace{-7pt}
\begin{equation}\label{eq:refl}
\Refl \left( \incidentdir, \outgoingdir \right) =  \alpha_d \, \Refl_d \left( \incidentdir, \outgoingdir \right) \,\, + \,\, \underbrace{\alpha_s \, \Refl_s \left( \incidentdir, \outgoingdir \right)}_{\approx 0}.
\vspace{-4pt}
\end{equation}
In contrast to recent work~\cite{chen2019steady,lindell2019acoustic}, we cannot rely on the specular component $\Refl_s$, as for large standoff distances, the relay walls are too small to capture the specular reflection.
Indeed, completely specular facet surfaces are used as ``stealth'' technology to hide targets~\cite{lynch2004introduction}. As retroreflective radar surfaces are extremely rare in nature~\cite{richards2010principles}, the diffuse part $\Refl_d$ dominates $\Refl$.
Note that $\att(\voxelpos) \Refl \left( \voxelpos - \wallpos, \wallpos - \voxelpos \right)$ in Eq.~\eqref{eq:transport_model} is known as the intrinsic radar albedo, describing backscatter properties, \ie, the {radar cross section}~\cite{sarabandi1997modeling}.

\vspace{0.8em}\noindent\textbf{Range Measurement}
Assuming an emitter and detector position $\rx = \tx = \wallpos$ and a static single target $\xi$ at distance $r = \|\pixelpos - \voxelpos\|$ with roundtime reflection $\tau_\xi = \nicefrac{2r}{\clight}$,  Eq.~\eqref{eq:transport_model} becomes a single sinusoid 
\vspace{-4pt}
\begin{align}
\quad & s_\xi(t) = \alpha_\xi g( t - \tau_\xi), \label{eq:single_reflector}
\vspace{-6pt}
\end{align}
where $\alpha_\xi$ describes here the accumulated attenuation along the reflected path. FMCW radars mix the received signal $s_\xi$ with the emitted signal $g$, resulting in a signal $p_\xi$ consisting of the sum and a difference of frequencies. The sum is omitted due to low-pass filtering in the mixing circuits, \ie:
\vspace{-1pt}
\begin{equation}  \label{eq:mixed_signal}
\! p_\xi(t) = s_\xi(t) \cdot g(t)  \approx \frac{\alpha_\xi}{2} \cos \left(2\pi \fbeat t + 4\pi \frac{f_c r}{\clight}  \right).
\end{equation}
\vspace{-1pt}
The remaining difference due to the time difference between transmitted and received chirp, see Supplemental Material, results in a frequency shift with beat frequency
\vspace{-4pt}
\begin{equation}
\fbeat = \frac{B}{\risetime}\frac{2r}{\clight}, \qquad \text{and} \qquad r = \clight \, \frac{\fbeat \risetime}{2 B}.
\label{eq:beat_frequency}
\vspace{-3pt}
\end{equation}
The range can be estimated from this beat note according to Eq.~\eqref{eq:beat_frequency}. 
To this end, FMCW radar systems perform a Fourier analysis, where multiple targets with different path lengths (Eq.~\eqref{eq:transport_model}) appear in different beat frequency bins.

\vspace{0.8em}\noindent\textbf{Doppler Velocity Estimation}
For the case when the object is moving, radial movement $\radvel$ along the reflection path results in an additional Doppler frequency shift in the received signal
\vspace{-3pt}
\begin{equation}
\fdoppler = 2\cdot \frac{\radvel}{\lambda}.
\vspace{-3pt}
\end{equation}
To avoid ambiguity between a frequency shift due to round-trip travel opposed to relative movement, the ramp slope $B/\risetime$ is chosen high, so that Doppler shifts are negligible in Eq.~\eqref{eq:beat_frequency}. Instead, this information is recovered by observing the phase shift $\theta$ in the signals between two consecutive chirps with spacing $\chirptime$, see Fig.~\ref{fig:supp_chirp}, that is
\vspace{-3pt}
\begin{equation}
\radvel = \frac{\lambda \cdot \theta}{4\pi \cdot \chirptime} = \Vect{v} \cdot \frac{\virtualpos - \pixelpos}{\| \virtualpos - \pixelpos \|}.
\vspace{-3pt}
\end{equation}
This velocity estimate is the radial velocity, see Fig.~\ref{fig:supp_array}.
Akin to the range estimation, the phase shift $\theta$ (and velocity) is also estimated by a Fourier analysis, but applied on the phasors of $N$ sequential chirps for each range bin separately.

\vspace{0.8em}\noindent\textbf{Incident Angle Estimation}
To resolve incident radiation directionally, radars rely on an array of antennas. Under a far field assumption, \ie, $r \gg \lambda$, for a single transmitter and target, the incident signal is a plane wave. The incident angle $\phi$ of this waveform causes a delay of arrival $d \sin(\phi) / \clight$ between the two consecutive antennas with distance $d$, see Fig.~\ref{fig:supp_array}, resulting in a phase shift $\phaseshift = 2\pi d \sin(\phi)/\lambda$. Hence, we can estimate 
\vspace{-3pt}
\begin{equation}\label{eq:angle}
\phi = \arcsin \frac{ \phaseshift \lambda}{2\pi d}.
\end{equation} 
For this angle estimation, a single transmitter antenna illuminates and all receiver antennas listen. A frequency analysis on the sequence of phasors corresponding to peaks in the 2D range-velocity spectrum assigns angles, resulting in a 3D range-velocity-angle data cube. 

\vspace{-3pt}
\subsection{Sensor Post-Processing} \label{sec:postproc}
The resulting raw 3D measurement cube contains $1024 \times 512 \times 64$ bins for range, angle, and velocity, respectively. For low-reflectance scenes, typical noise, and clutter, tens of millions of non-zero reflection points can be measured. To tackle such measurement rates in real-time, we implement a constant false alarm rate filter to detect high RCS values $\tilde{\sigma}$ 
following~\cite{Rohling1983}. 
We retrieve a radar point cloud $\pointcloud$ with less than $10^4$ points, allowing for efficient inference:
\vspace{-4pt}
\begin{equation}
\pointcloud = \big\{ (\tilde{\yaw},\tilde{r},\tilde{\radvel},\tilde{\sigma})_i \; | \; 1 \leq i \leq R \big\} \text{ with } R < 10^4.
\label{eq:radar_pointcloud}
\vspace{-4pt}
\end{equation}
See Supplemental Material for details on post-processing.

\section{Joint NLOS Detection and Tracking}
In this section, we propose a neural network for the detection and tracking of hidden objects from radar data. 
\begin{figure*}[t]
\vspace{-0.5cm}
    \centering
    \includegraphics[width=0.97\textwidth]{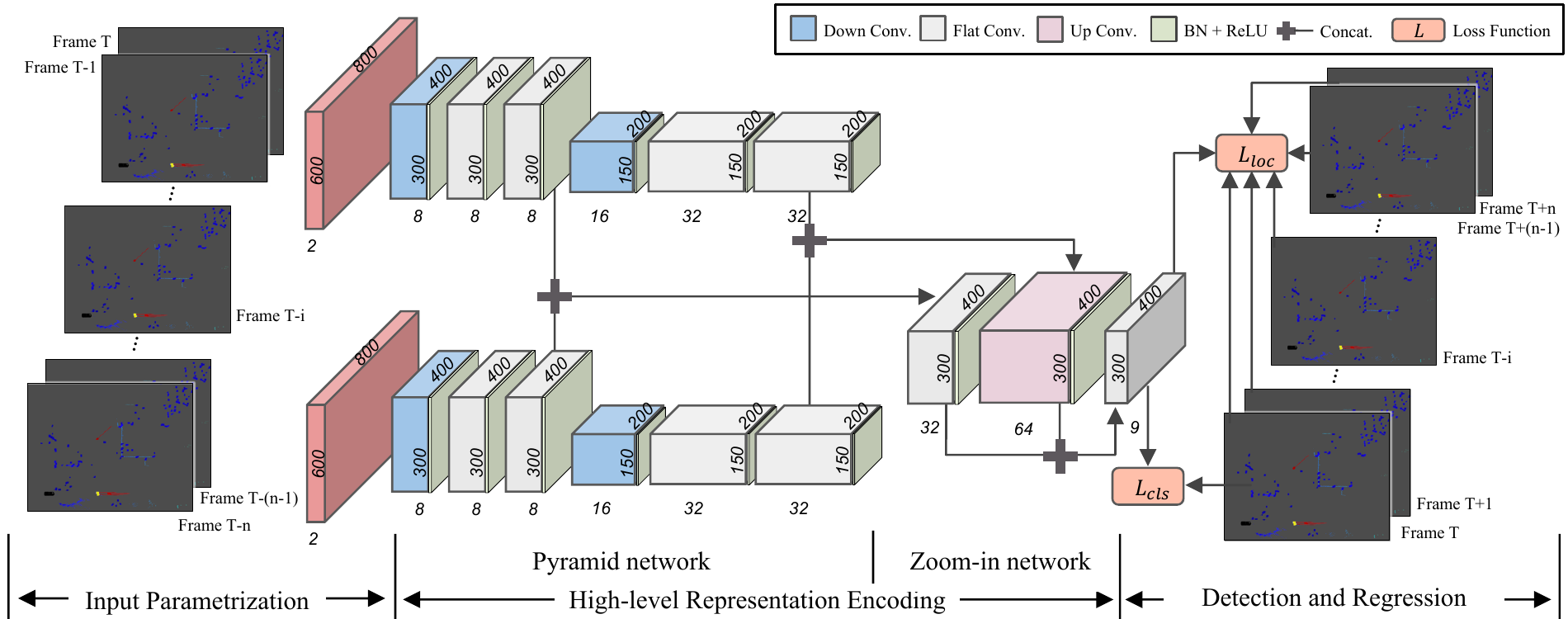}
    \vspace*{-0.05cm}
    \caption{NLOS detection and tracking architecture. The network accepts the current frame $T$ and the past $n$ radar point cloud data as input, and outputs predictions for frame $T$ and the following $n$ frames. The features are downsampled twice in the pyramid network, and then upsampled and concatenated by the zoom-in network. We merge the features from different frames at both levels to encode high-level representation and fuse temporal information. }
    \label{fig:Networks}
    \vspace{-0.6cm}
\end{figure*}

\subsection{Non-Line-of-Sight Detection}
%
\begin{figure}[t!]
	\vspace{-0.2cm}
	\centering
	\includegraphics[width=0.77\linewidth]{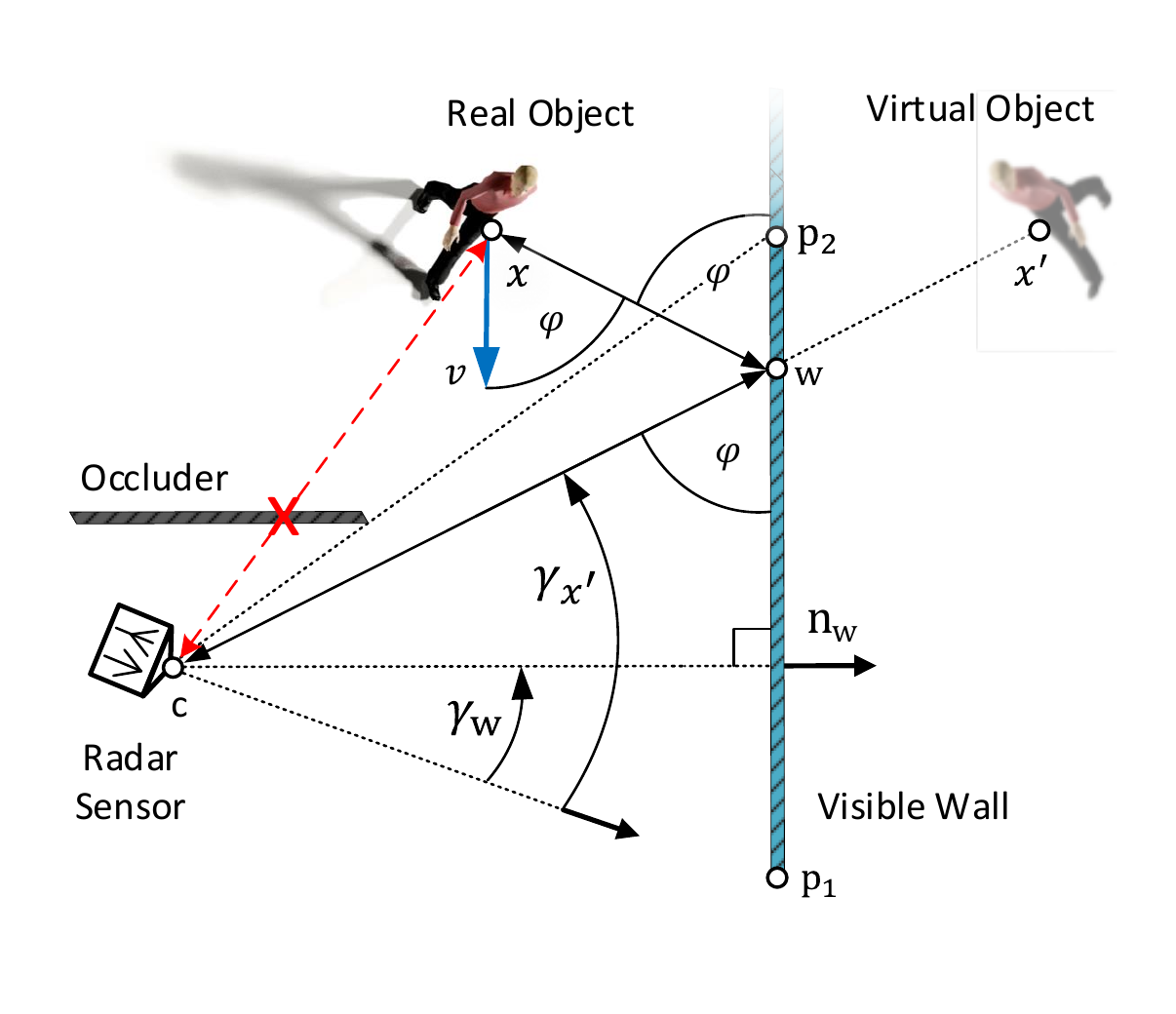}
	\vspace{-12pt}
	\captionsetup{skip=-2pt}
	\caption{NLOS geometry and velocity estimation from indirect specular wall reflections. The hidden velocity $v$ can be reconstructed from the radial velocity $v_r$ by assuming that the road user moves parallel to the wall, \ie, on a road.}
	\label{fig:NLOS_scheme}
	\vspace{-15pt}
\end{figure}
The detection task is to estimate oriented 2D boxes for pedestrians and cyclists, given a Bird's-eye-view (BEV) point cloud $\pointcloud$ as input. The overall detection pipeline consists of three main stages: (1) input parameterization that converts a BEV point cloud into a sparse pseudo-image; (2) high-level representation encoding from the pseudo-image using a 2D convolutional backbone; and (3) 2D bounding box regression and detection with a detection head.

\vspace{0.8em}\noindent\textbf{Input Parameterization}\label{detect}
We denote $\Vect{u}$ as a $d$-dimensional ($d=4$) point in a raw radar point cloud $\pointcloud$ with coordinates $x$, $y$ (derived from the polar coordinates $\tilde{\yaw},\tilde{r}$), velocity $\tilde{\radvel}$, and amplitude $\tilde{\sigma}$. We use the logarithm of the amplitude to get an intensity measure $s=\log \tilde{\sigma}$. As a first step, the point cloud is discretized into an evenly spaced grid in the $x$-$y$ plane, resulting in a pseudo-image of size $(d-2, H, W)$ where $H$ and $W$ indicate the height and width of the grid.

\vspace{0.8em}\noindent\textbf{High-level Representation Encoding}
To efficiently encode high-level representations of the hidden detections, the backbone network contains two modules: a pyramid network and a zoom-in network. The pyramid network contains two consecutive stages to produce features at increasingly small spatial resolutions. Each stage downsamples its input feature map by a factor of two using three 2D convolutional layers. Next, a zoom-in network upsamples and concatenates the two feature maps from the pyramid network. This zoom-in network performs transposed 2D convolutions with different strides. As a result, both upsampled features have the same size and are then concatenated to form the final output.
All (transposed) convolutional layers use kernels of size 3 and are interlaced with BatchNorm and ReLU, see Supplemental Material for details.

\vspace{0.8em}\noindent\textbf{Detection Head}
The detection head follows the setup of Single Shot Detector (SSD)~\cite{lin2017focal} for 2D object detection. Specifically, each anchor predicts a 3-dimensional vector for classification (background / cyclist / pedestrian) and a 6-dimensional vector for bounding box regression (center, dimension, orientation, and velocity of the box).

\vspace{0.8em}\noindent\textbf{Relay Wall Estimation}
We use first-response pulsed lidar measurements of a separate front-facing lidar sensor to recover the geometry of the visible wall. Specifically, we found that detecting line segments in a binarized binned BEV is robust using \cite{von2012lsd}, where each bin with size \SI{0.01}{\meter} is binarized with a threshold of 1 detection per bin. We filter out segments with a length shorter than \SI{1}{\meter}, constraining the detected wall to smooth surfaces that our NLOS model holds for, see Supplemental Material.
Each segment is represented by its endpoints $\Vect{p}_1$ and $\Vect{p}_2$, cf. Fig.~\ref{fig:NLOS_scheme}.

\vspace{0.8em}\noindent\textbf{Third-Bounce Geometry Estimation}
Next, we derive the real location $\voxelpos$ of a third-bounce or virtual detection $\virtualpos$, for reference see Fig.~\ref{fig:supp_array} and Fig.~\ref{fig:NLOS_scheme}.
In order to decide whether a point is a virtual detection, we first derive its intersection $\wallpos$ with the relay wall $\wall = \Vect{p}_2 - \Vect{p}_1$, that is
\vspace{-7pt}
\begin{equation}
\wallpos = \rx + \frac{(\Vect{p}_1 - \rx) \cross \wall}{(\virtualpos - \rx) \cross \wall} (\virtualpos - \rx),
\vspace{-6pt}
\end{equation}
where $\cross$ is the 2D cross product $\Vect{a} \cross \Vect{b} = a_1b_2 - a_2b_1$.
For a detection $\virtualpos$ to be a third-bounce detection, we have two criteria.
First, $\virtualpos$ and the receiver $\rx$ must be on opposite sides of the relay wall.
We define the normal of the relay wall $\wallnormal$ as pointing away from the receiver $\rx$.
Second, the intersection $\wallpos$ must be between $\Vect{p}_1$ and $\Vect{p}_2$, both expressed as
\vspace{-4pt}
\begin{equation}
\label{eq:is-virtual}
\begin{aligned}
\wallnormal \cdot (\virtualpos - \Vect{p}_1) \geq 0 & \land 	\norm{\wallpos - \Vect{p}_1} \leq \norm{\wall} \\ & \land \norm{\wallpos - \Vect{p}_2} \leq \norm{\wall}.
\end{aligned}
\vspace{-4pt}
\end{equation}
The first term is the signed distance, indicating whether $\virtualpos$ and $\rx$ are on opposite sides of the wall and the other terms determine whether $\wallpos$ lies between $\Vect{p}_1$ and $\Vect{p}_2$.
If Eq.~\eqref{eq:is-virtual} is true, \ie, $\virtualpos$ is a third-bounce detection, we reconstruct the original point $\voxelpos$ as
\vspace{-4pt}
\begin{equation}
	\voxelpos
	= \frac{\big(\wallpos - \rx
	- 2\left(\wallnormal\cdot\left(\wallpos - \rx\right) \right)\wallnormal\big) \norm{\wallpos - \virtualpos}}{\norm{\wallpos - \rx}}.
\end{equation}

\vspace{-10pt}
\vspace{0.8em}\noindent\textbf{Third-Bounce Velocity Estimation}
After recovering $\voxelpos$, we estimate the real velocity vector $\Vect{v}$ under the assumption that the real velocity is parallel to the relay wall, see Fig.~\ref{fig:NLOS_scheme}.
Specifically, it is
\vspace{-9pt}
\begin{equation}\label{eq:v-direction}
	\Vect{v} = \norm{\Vect{v}} \sign(v_r) \cdot \sign(\gamma_\virtualpos - \gamma_\wallpos) \frac{\wall}{\norm{\wall}}.
\vspace{-6pt}
\end{equation}
Here, $\gamma_\virtualpos$ and $\gamma_\wallpos$ are the angles of $\virtualpos - \rx$ and $\wallnormal$ relative to the sensor's coordinate system, respectively.
In Eq.~\eqref{eq:v-direction}, the sign of $\radvel$ distinguishes approaching and departing hidden object targets, while $\sign(\gamma_\virtualpos - \gamma_\wallpos)$ determines the object's allocation to the left or right half-plane with respect to the normal $\wallnormal$.
By convention, we define that $\wall$ is rotated $\frac{\pi}{2}$ counterclockwise relative to $\wallnormal$.
Using the measured radial velocity $\radvel= \norm{\Vect{v}} \cdot \abs{\cos{\varphi}}$, we get 
\vspace{-7pt}
\begin{equation}
\Vect{v} =  \sign(v_r) \cdot \sign(\gamma_\virtualpos - \gamma_\wallpos) \cdot \frac{\abs{v_r}}{\abs{\cos{\varphi}}} \cdot \frac{\wall}{\norm{\wall}},
\vspace{-4pt}
\end{equation}
with $\varphi$ being the angle between $\virtualpos - \rx$ and $\Vect{v}$, cf. Fig. \ref{fig:NLOS_scheme}.
See the Supplemental Material for detailed derivations.

\subsection{Non-Line-of-Sight Doppler Tracking}

Our model jointly learns tracking with future frame prediction, inspired by Luo et al.~\cite{luo2018fast}. At each timestamp, current and its $n$ preceding frames form the input, and predictions are for the current plus the following $n$ future frames.

One of the main challenges is to fuse temporal information. A straightforward solution is to add another dimension and perform 3D convolutions over space and time. However, this approach is not memory-efficient and computationally expensive given the sparsity of the data. Alternatives can be early or late fusion as discussed in \cite{luo2018fast}. Both fusion schemes first process each frame individually, and then start to fuse all frames together. 

Instead of such one-time fusion, our approach leverages the multi-scale backbone and performs fusion at different levels. Specifically, we first perform separate input parameterization and high-level representation encoding for each frame as described in Sec.~\ref{detect}. After the two stages of the pyramid network, we concatenate the $n+1$ feature maps along the channel dimension for each stage. This results in two feature maps of sizes $\left((n+1)C_1, \frac{H}{2}, \frac{W}{2}\right)$ and $\left((n+1)C_2, \frac{H}{4}, \frac{W}{4}\right)$, which are then concatenated as inputs to the two upsampling modules of the zoom-in network, respectively. The rest of the model is the same as before. By aggregating temporal information across $n+1$ frames at different scales, the model is allowed to capture both low-level per-frame details and high-level motion features. We refer to Fig.~\ref{fig:Networks} for an illustration of our architecture.

\subsection{Loss Functions}
Our overall objective function contains a localization term and a classification term 
\vspace{-6pt}
\begin{equation}
L=\alpha L_{loc} + \beta L_{cls}.
\vspace{-4pt}
\end{equation}
The localization loss is a sum of the localization loss for the current frame $T$ and $n$ frames into the future:
\vspace{-6pt}
\begin{equation}
L_{loc} =\sum_{t=T}^{T+n} L_{{loc}_t} \;\;\; \text{with} \;\;\; L_{{loc}_t} = \!\!\!\!\!\! \sum_{u\in\{x, y, w, l, \theta,v\}} \hspace{-.7cm} \alpha_u |\Delta u|,
\vspace{-4pt}
\end{equation}
where $\Delta u$ is the localization regression residual between ground truth ($gt$) and anchors ($a$) defined by $(x,y,w,l,\theta,v)$:
\vspace{-1pt}
\begin{small}
\begin{align}
\Delta x = x^{gt} - x^{a}, \quad \Delta y = y^{gt} - y^{a}, \quad \Delta v = v^{gt} - v^{a}, \vspace{-6pt} \nonumber\\
\Delta w =\log\frac{w^{gt}}{w^{a}}, \quad \Delta l =\log\frac{l^{gt}}{l^{a}}, \quad \Delta \theta = \sin(\theta^{gt}-\theta^{a}).\vspace{-4pt}
\end{align}
\end{small}
We do not distinguish the front and back of the object, therefore all $\theta$'s are within the range $[-\frac{\pi}{2},\frac{\pi}{2})$.

\section{Data Acquisition and Training}
\vspace{-4mm}
\begin{figure}[!t]
	\vspace{-10pt}
    \centering
		\includegraphics[width=\linewidth]{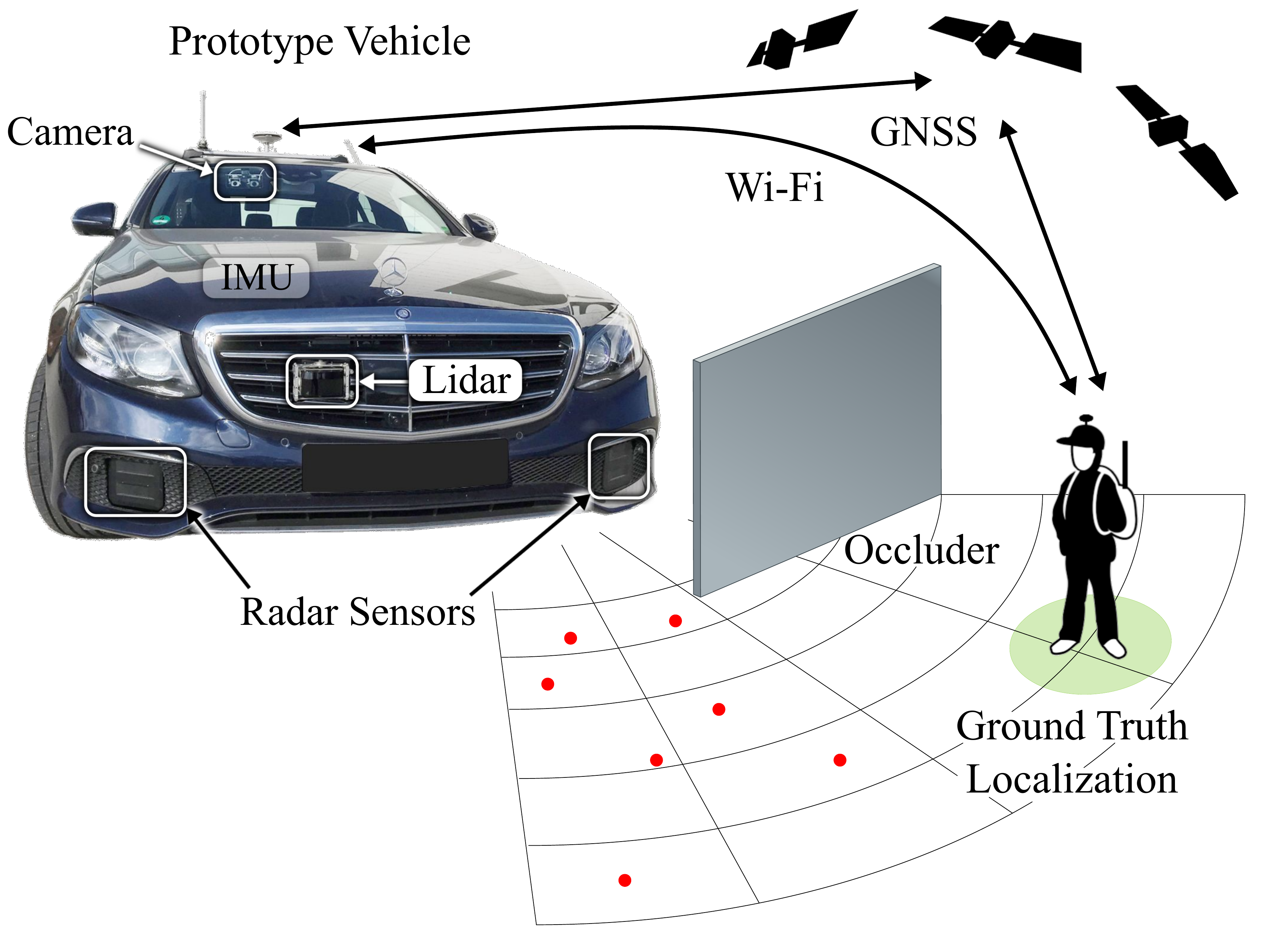}
	\captionsetup{skip=-2pt}
    \caption{Prototype vehicle with measurement setup (top left) and automated ground-truth localization system (right). To acquire training data in an automated fashion, we use GNSS and IMU for a full pose estimation of ego-vehicle and the hidden vulnerable road users.}
    \label{fig:schematic_pic2}
    \vspace{-12pt}
\end{figure}
\vspace{0.8em}\noindent\textbf{Prototype Vehicle Setup}
The observation vehicle prototype is shown in Fig.~\ref{fig:schematic_pic2}.
We use experimental FMCW radar prototypes, mounted in the front bumper, with frequency band \SIrange{76}{77}{\giga\hertz} and chirp sequence modulation, see Sec.~\ref{sec:image_formation}. We use a mid-range configuration with \SI{153}{\meter} maximum range and FoV of \SI{140}{\degree}, \ie, for urban scenarios or intersections. A single measurement takes \SI{22.6}{\milli\second}, with a resolution of \SI{0.15}{\meter}, \SI{1.8}{\degree}, and \SI{0.087}{\meter\per\second}. Similar sensors are available as development kits for a few hundred USD, \eg Texas Instruments AWR1642BOOST; the mass-produced version costing a small fraction. The radar sensors are complemented by an experimental 4-layer scanning lidar with \SI{0.25}{\degree} and \SI{0.8}{\degree} resolution in azimuth and elevation. With a wide FoV of \SI{145}{\degree}, a single sensor installed in the radiator grill suffices for our experiments. We use a GeneSys ADMA-G PRO localization system consisting of a combined global navigation satellite system (GNSS) receiver and an inertial measurement unit (IMU) to track ego-pose using an internal Kalman filter. The system has an accuracy of up to \SI{0.8}{\centi\meter} and \SI{0.01}{\meter\per\second} for the position and velocity. For documentation purposes, we use a single AXIS F1015 camera with \SI{97}{\degree} FoV behind the test vehicle's windshield.
See Supplemental Material for details on all sensors along with required coordinate system transforms.

\vspace{0.8em}\noindent\textbf{Automated Ground-Truth Estimation}
Unfortunately, humans are not accustomed to annotating radar measurements, and NLOS annotations are even more challenging.
We tackle this problem by adopting a variant of the tracking device from~\cite{Scheiner2019ICMIM}. We equip vulnerable road users, \ie, occluded pedestrians or bicyclists, with a hand-held GeneSys ADMA-Slim tracking module synced with the capture vehicle via Wi-Fi, see Fig.~\ref{fig:schematic_pic2}. 
In contrast to~\cite{Scheiner2019ICMIM} we do not purely rely on GNSS, but also use the IMU for pose estimation of the hidden object, see Supplemental Material.

\vspace{0.4em}\noindent\textbf{Training and Validation Data Set}~\label{sec:dataset}
\begin{figure}[t!] 
\vspace{-10pt}
  \centering
	\begin{subfigure}{.49\linewidth}
	\includegraphics{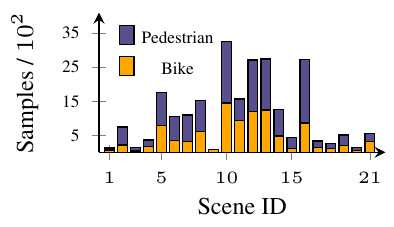}
	\captionsetup{skip=1pt, margin=.5cm, font=footnotesize}
	\caption{NLOS detection sample distribution over scenarios.}
	\label{fig:data_stats_b2}
	\vspace{3pt}
	\end{subfigure}
	\begin{subfigure}{.49\linewidth}
	\includegraphics{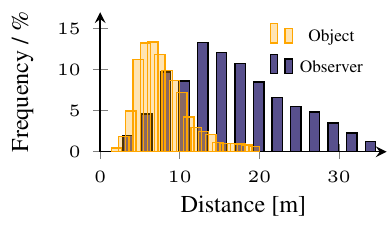}
	\captionsetup{skip=1pt, margin=.5cm, font=footnotesize}
	\caption{Distance of hidden object and observer to wall.}
	\label{fig:data_stats_c2}
	\vspace{3pt}
	\end{subfigure}
	\input{figures/dataset_overview_mainpaper3}
	\captionsetup{skip=2pt}
	\caption{NLOS training and evaluation data set for large outdoor scenarios. Top: Data set statistics (a), and hidden object and observer distances (b) to the relay wall. Bottom: Camera images including the (later on) hidden object. }
	\label{fig:data_stats_scene_examples}
	\vspace{-14pt}
\end{figure}
We capture a total of $100$ sequences in-the-wild automotive scenes with $21$ different scenarios, \ie, we repeat scenarios with different NLOS trajectories multiple times.
The wide range of relay walls appearing in this dataset is shown in Fig.~\ref{fig:data_stats_scene_examples} and includes plastered walls of residential and industry buildings, marble garden walls, a guard rail, several parked cars, garages, a warehouse wall, and a concrete curbstone.
The dataset is equally distributed among hidden pedestrians and cyclists, and adds up to over $32$ million radar points, see Supplemental Material.
We opt for these two kinds of challenging road users, as bigger, faster, and more electrically conductive objects such as cars are much easier to detect for automotive radar systems.
We split the dataset into non-overlapping training and validation sets, where the validation set consists of four scenes with $20$ sequences and $3063$ frames.

\vspace{-1mm}
\section{Assessment}\label{sec:eval}
\vspace{-1mm}

\vspace{0.1em}\noindent\textbf{Evaluation Setting and Metrics}
For both, training and validation, the region of interest is a large area of \SI{60}{\meter} $\times$ \SI{80}{\meter}. We use resolution \SI{0.1}{\meter} to discretize both $x,z$ axes into a $600\times 800$ grid. We assign each ground truth box to its highest overlapping predicted box for training. The hidden classification and localization performance are evaluated with Average Precision (AP) and average distance between the predicted and ground truth box centers, respectively.
\begin{figure*}[t]
    \centering
		\vspace*{-0.5cm}
    \includegraphics[width=0.98\linewidth]{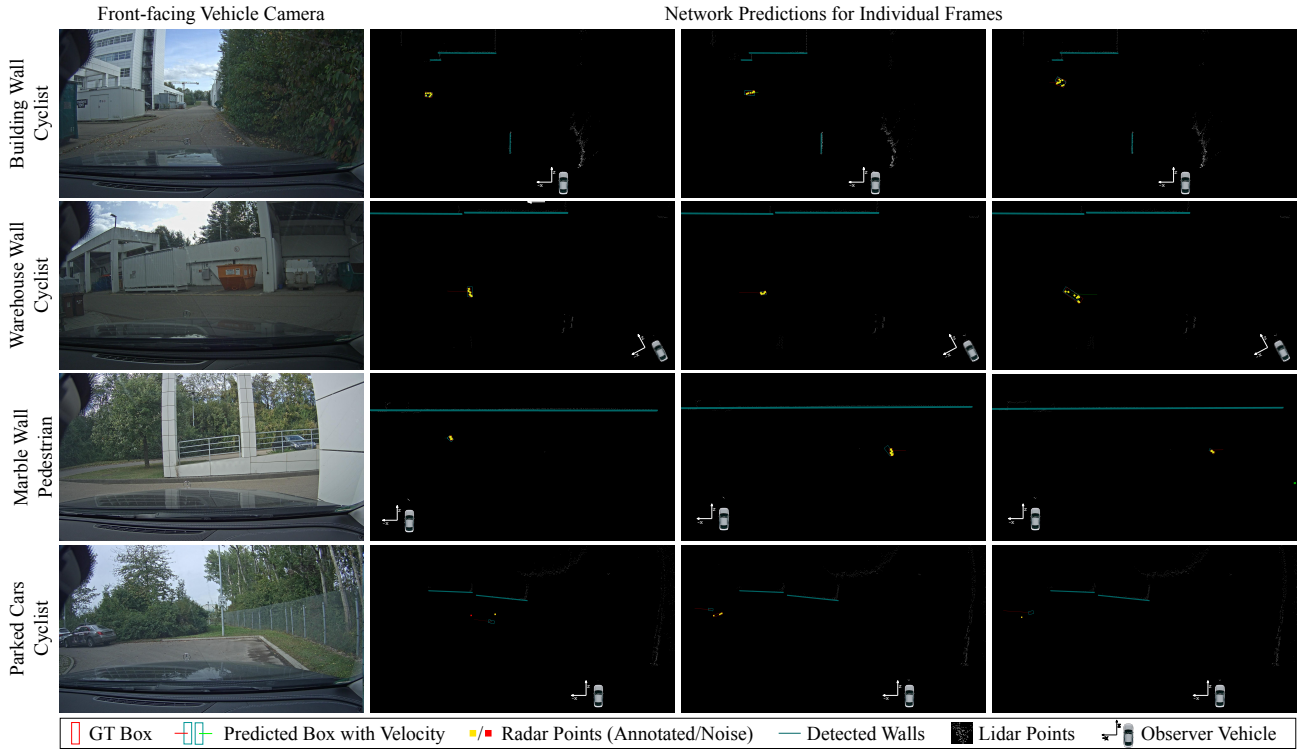}
    \vspace*{-0.2cm}
    \caption{Joint detection and tracking results for automotive scenes with different relay wall type and object class in each row. The first column shows the observer vehicle front-facing camera view. The next three columns plot BEV radar and lidar point clouds together with bounding box ground truth and predictions. NLOS velocity is plotted as line segment from the predicted box center: red and green corresponds to moving towards and away from the vehicle.}
    \label{fig:Results_detect_track}
    \vspace{-0.2cm}
\end{figure*}

\begin{figure}[t]
	\vspace{0.11cm}
	\centering
	\begin{subfigure}{.5\linewidth}
		\centering
		\includegraphics[width=1\linewidth]{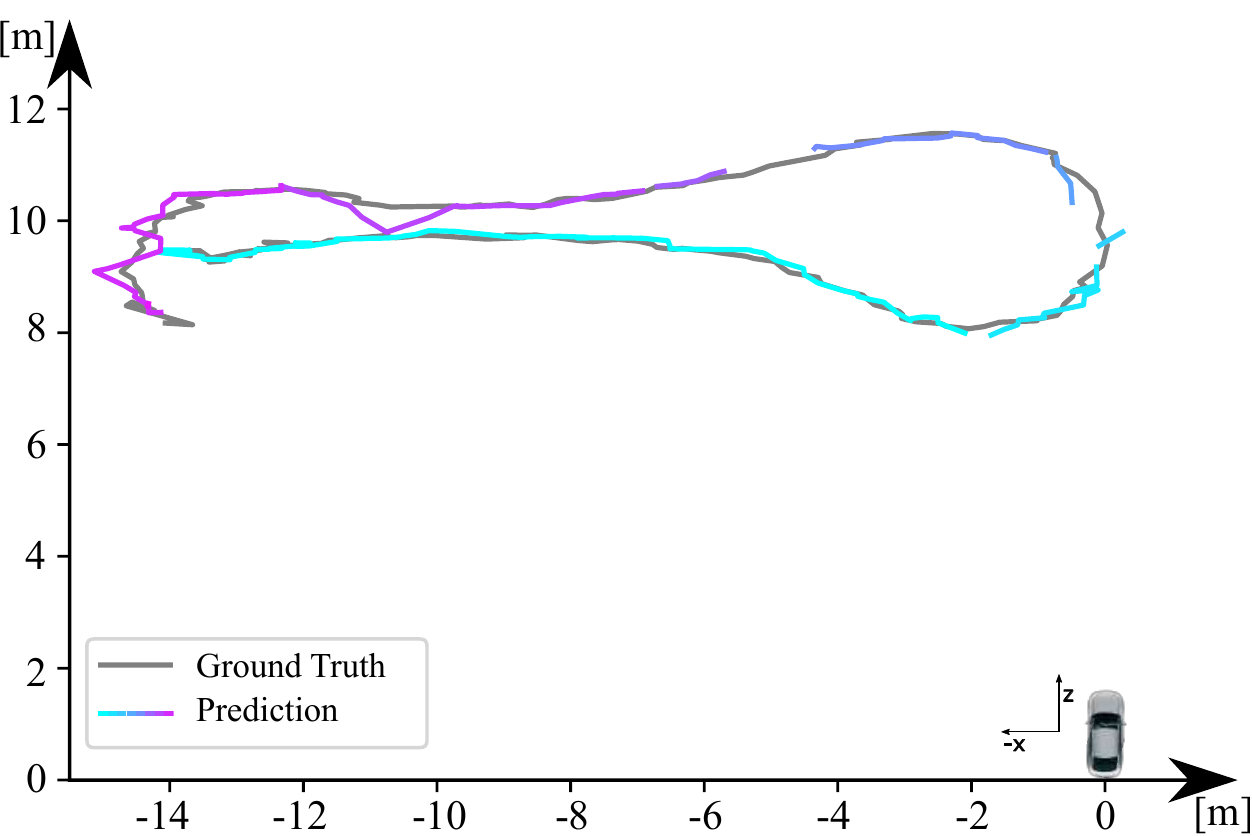}
		\captionsetup{skip=3pt, font=small}
		\caption{Cyclist}
	\end{subfigure}%
	\begin{subfigure}{.5\linewidth}
		\centering
		\includegraphics[width=1\linewidth]{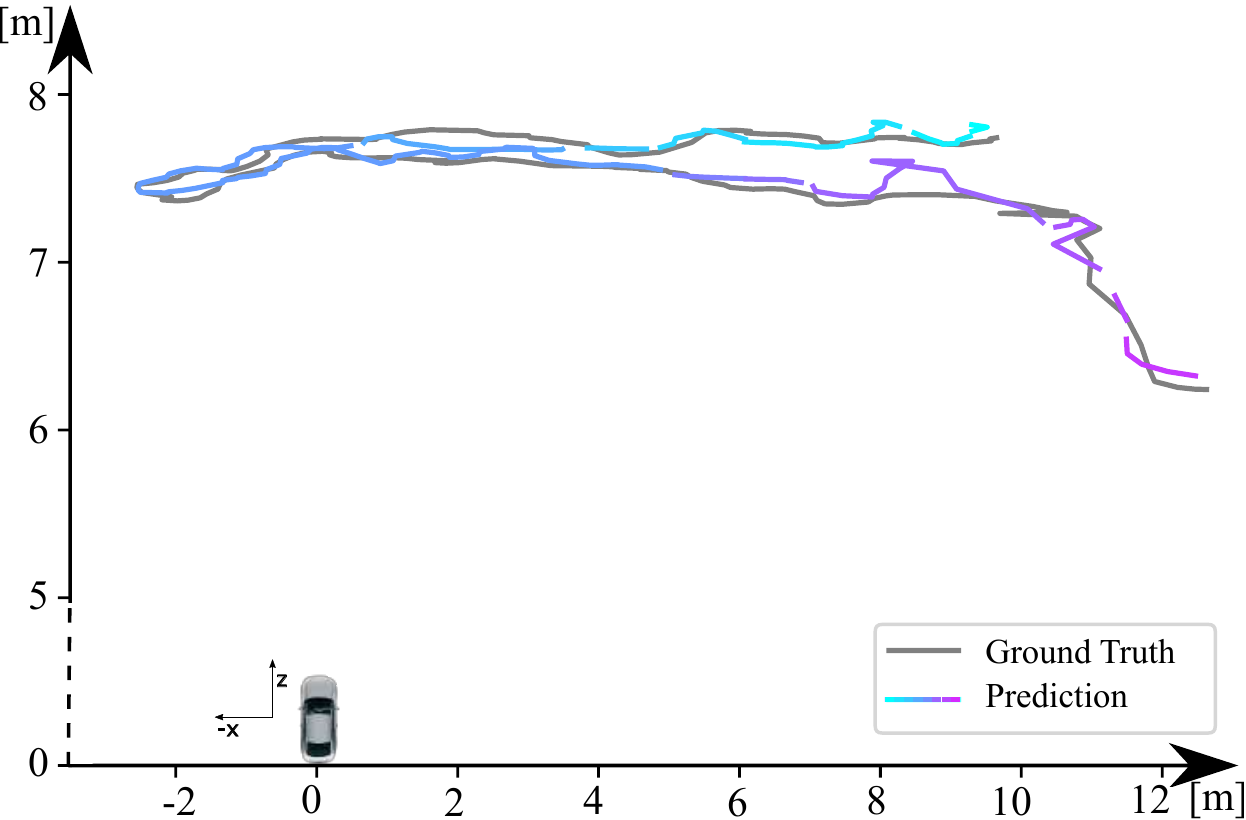}
		\captionsetup{skip=3pt, font=small}
		\caption{Pedestrian}
	\end{subfigure}
	\captionsetup{skip=1pt}
	\caption{Tracking trajectories for two NLOS scenes. The predictions consist of segments, with each corresponding to a different tracking ID  visualized in different colors.}
	\label{fig:Results_track}
	\vspace{-0.5cm}
\end{figure}

\vspace{1.2em}\noindent\textbf{Qualitative Validation}
Fig.~\ref{fig:Results_detect_track} shows results for realistic automotive scenarios with different wall types.
Note that the size of ground truth bounding box varies due to the characteristics of radar data. The third row shows a scenario where no more than three detected points are measured for the hidden object, and the model has to rely on velocity and orientation of these sparse points to make a decision on box and class prediction. Despite such noise, we do observe that the model outputs stable predictions. As illustrated in the fourth row, predicted boxes are very consistent in size and orientation across frames despite the extreme radar detection sparsity. The first frame in the fourth row shows a detection where a hidden object became visible by lidar but not radar. Note that all other scenes have occluder geometries visible in the lidar measurements.
For rare cases where the ground truth information is imperfect due to jitter of the ground truth acquisition system, we can reason about sequences of frames instead of a single one.
While the predicted box seemingly does not match the ground truth well in this particular frame, it is, in fact, detected correctly, validating the proposed joint detection and tracking approach.
Fig.~\ref{fig:Results_track} shows qualitative tracking trajectories for two different scenes. The model is able to track an object only with occasional incorrect ID switch. 

\vspace{0.2em}\noindent\textbf{Quantitative Detection Results}
\begin{table}
\vspace{-0.4cm}
\begin{center}
\addtolength{\tabcolsep}{-4.4pt}
\footnotesize
\resizebox{\linewidth}{!}{
\begin{tabular}{lccccccccc}
\noalign{\hrule height 1pt}

Class~~~~~&\multicolumn{3}{c}{Cyclist}&\multicolumn{3}{c}{Pedestrian}&\multicolumn{3}{c}{Object }\\
\cmidrule(lr){2-4}\cmidrule(lr){5-7}\cmidrule(lr){8-10}
AP \quad&${@}$0.5&${@}$0.25&${@}$0.1&${@}$0.5&${@}$0.25&${@}$0.1&${@}$0.5&${@}$0.25&${@}$0.1\\
\midrule
Ours&\textbf{29.35}&\textbf{56.43}&\textbf{62.40}&\textbf{44.74}&\textbf{62.19}&\textbf{68.15}&\textbf{41.36}&\textbf{66.34}&\textbf{75.41}\\
SSD~\cite{lin2017focal}\footnotemark\addtocounter{footnote}{-1}\addtocounter{Hfootnote}{-1}&10.07&37.87&51.50&27.19&49.24&56.24&19.87&46.29&60.98\\
PointPillars~\cite{lang2019pointpillars}\footnotemark &2.02&15.02&28.00&7.83&22.16&26.76&9.61&45.69&58.68\\
\bottomrule
\end{tabular}}
\end{center}
\captionsetup{skip=-8pt}
\caption{Detection classification (AP) comparison. We compare our model to an SSD detector and the PointPillars~\cite{lang2019pointpillars}, details in Supplementary Material.}
\label{tab: detect-classification}
    \vspace{-0.2cm}
\end{table}
\footnotetext{Trained with proposed third-bounce geometry and velocity estimates.}
\begin{table}
\begin{center}
\addtolength{\tabcolsep}{-4.4pt}
\footnotesize
\begin{tabular}{lcc}
\toprule
\multicolumn{3}{c}{\textbf{Localization}}\\
\multicolumn{3}{c}{(Box Center Distance)}\\
\cmidrule(lr){1-3}
Model&MAE&MSE\\
\midrule
Tracking (w. $v$)&0.12&0.03\\
Tracking (w/o. $v$)\footnotemark\addtocounter{footnote}{-1}\addtocounter{Hfootnote}{-1} &0.13&0.04\\
\bottomrule
\end{tabular}
\quad
\begin{tabular}{lccc}
\toprule
Model&Visibility&MOTA&MOTP\\ 
\midrule
Tracking&NLOS&0.58&0.93 \\
(w. $v$)&LOS&0.85&0.91 \\
\midrule
Tracking&NLOS&0.52&0.94 \\
(w/o. $v$)\footnotemark&LOS&0.81&0.90 \\
\bottomrule
\end{tabular}
\end{center}
\captionsetup{skip=-8pt}
\caption{Localization and tracking performance on NLOS and LOS data, with MAE and MSE in meters. Velocity prediction (and supervision) indicated by $v$.}
\label{tab: tracking}
    \vspace{-0.4cm}
\end{table}
\footnotetext{Input is velocity-based pre-processed data, see Supplemental Material.}
We report AP at IoU thresholds $0.1, 0.25$ and $0.5$ for cyclist/pedestrian detection in Tab.~\ref{tab: detect-classification}. We also list the mean AP of predicting object/non-object by merging cyclist/pedestrian labels. We compare our model to a simplified SSD~\cite{lin2017focal} and the PointPillars~\cite{lang2019pointpillars} for BEV point cloud detection, see Supplemental Material. Since most bounding boxes in our collected data are challenging small boxes with sizes smaller than \SI{0.5}{\meter} $\times$ \SI{0.5}{\meter}, a very small offset may significantly affect the detection performance at a high IoU threshold. However, in practice, a positive detection with an IoU as small as 0.1 is still a valid detection for collision warning applications. Combined with the high localization accuracy, see Tab.~\ref{tab: tracking} (left), the proposed approach allows for long-range detection and tracking of hidden object in automotive scenarios, even for small road users as pedestrians and bicycles.

\vspace{0.2em}\noindent\textbf{Quantitative Tracking Results}
Tab.~\ref{tab: tracking} lists the localization and tracking performance of the proposed approach. Relying on multiple frames and measured Doppler velocity estimates, the proposed method achieves high localization accuracy of ~\SI{0.1}{\meter} in MAE despite measurement clutter and small diffuse cross section of the hidden pedestrian and bicycle objects. We evaluate the tracking performance on NLOS and visible line-of-sight (LOS) frames separately in Tab.~\ref{tab: tracking}. For challenging NLOS data, while the number of unmatched objects (Multiple Object Tracking Accuracy -- MOTA) increases, the model is still able to precisely locate most of the matched objects (Multiple Object Tracking Precision -- MOTP). These results validate the proposed joint NLOS detector and tracker for collision avoidance applications.  Tab. \ref{tab: tracking} also compares models with and without velocity supervision, showing that velocity supervision improves both localization and tracking accuracy.

\section{Conclusion}
In this work, we introduce a novel method for joint non-line-of-sight detection and tracking of occluded objects using automotive Doppler radar.
Learning detection and tracking end-to-end from a realistic NLOS automotive radar data set, we validate that the proposed approach allows for collision warning for pedestrians and cyclists in real-world autonomous driving scenarios -- before seeing them with existing direct line-of-sight sensors. In the future, detection from higher-order bounces, and joint optical and radar NLOS could be exciting next steps.

\section*{Acknowledgment}
This research received funding from the European Union under the H2020 ECSEL program as part of the DENSE project, contract number 692449.

\clearpage
{\small
\bibliographystyle{ieee}
\bibliography{bib}
}

\end{document}